\documentclass{article}

\usepackage[final]{nips_2018}

\usepackage[utf8]{inputenc} 
\usepackage[T1]{fontenc}    
\usepackage{hyperref}       
\usepackage{url}            
\usepackage{booktabs}       
\usepackage{amsfonts}       
\usepackage{nicefrac}       
\usepackage{microtype}      
\usepackage{graphicx}
\usepackage{commath, bm, bbm, mathtools, amssymb}
\usepackage{amsmath}
\usepackage{mathtools}
\usepackage{amsfonts}
\usepackage{amssymb}
\usepackage{fancyhdr}
\usepackage{xcolor}
\usepackage{booktabs}
\usepackage{float}
\usepackage{graphicx}
\usepackage{caption}
\usepackage{verbatim}

\newcommand{\expt}[0]{\mathbb{E}}
\newcommand{\decoder}{\mathcal{D}}
\newcommand{\encoder}{\mathcal{E}}
\newcommand{\data}{\mathbf{x}}

\title{Image Style Transfer and Content-Style Disentanglement}

\author{
  Sailun Xu \\
  School of Computer Science\\
  Carnegie Mellon University\\
  Pittsburgh, PA 15213 \\
  \texttt{sailunx@andrew.cmu.edu} \\
  \And
  Jiazhi Zhang \\
   School of Computer Science\\
  Carnegie Mellon University\\
  Pittsburgh, PA 15213 \\
  \texttt{jiazhiz@andrew.cmu.edu} \\
  \And
  Jiamei Liu \\
  School of Computer Science\\
  Carnegie Mellon University\\
  Pittsburgh, PA 15213 \\
  \texttt{jiameil@andrew.cmu.edu} 
}

\begin{document}

\maketitle
\begin{abstract}
    We propose a way of learning disentangled content-style representation of image, allowing us to extrapolate images to any style as well as interpolate between any pair of styles. By augmenting data set in a supervised setting and imposing triplet loss, we ensure the separation of information encoded by content and style representation. We also make use of cycle-consistency loss to guarantee that images could be reconstructed faithfully by their representation.
\end{abstract}

\section{Introduction}
Image transformation is a classic problem in computer vision. In recent years, topics of image style transfer and texture synthesis have been revitalized due to the emergence of Convolutional Neural Network (\cite{Johnson2016PerceptualLF}), which brought a new wave of performance increase to the task of image style transfer. In this project, we plan to learn a disentangled image representation consists of content and style using an augmented data set. We hope this disentangled formulation could allow us to easily interpolate images between any style and extrapolate to any unseen style by simply swapping the style encoding.

\section{Related Work}
\subsection{Neural Style Transfer}
Rendering the content information of an image in the style of a different given image has been a long-standing branch in image transformation. The difficulty is deep rooted in how to get a semantic representation of an image that well separates \texttt{content} from \texttt{style} according to human's conception. \citep{cnn_style_transfer} first applied CNN to this problem, but relied on a rather constrained and arbitrary content reconstruction and style reconstruction loss. Specifically, suppose that the feature map of a CNN at layer $l$ has shape $[C,H,W]$ where $C,H,W$ represent channel, height and width respectively, then content is taken as the feature map of a CNN's response at one or more layers $l$: $\mathcal{F}^l \in \mathbb{R}^{C, H, W}$ and style taken as the gram matrix of the feature map: $\mathcal{G}^l \in \mathbb{R}^{C \times C}$

\[ \mathcal{G}^l_{c_1,c_2} = \sum_{h,w} \mathcal{F}^l_{c_1, h, w}\mathcal{F}^l_{c_2, h, w} = \langle \mathcal{F}^l_{c_1}, \mathcal{F}^l_{c_2} \rangle \]

During the test time, they jointly minimize the content difference with the content-target image and style difference with the style-target image using L-BFGS, making the inference rather slow. Moreover, this method offers no intuitive information about the style code. 

Along with this line, some other methods have been proposed to accelerate the testing time. \cite{real-time-transfer} first used a co-match layer:
\[ \hat{\mathcal{Y}}^{i}=\Phi^{-1}\left[\Phi\left(\mathcal{F}^{i}\left(\textbf{x}_{c}\right)\right)^{T} W \mathcal{G}\left(\mathcal{F}^{i}\left(\textbf{x}_{s}\right)\right)\right]^{T}\] 

Where $\Phi$ is a reshaping operation to match dimension, $W$ is a trained parameter to automatically balance the trade-off between content feature $\mathcal{F}$ of the content target $x_c$ and the style feature $\mathcal{G}(\mathcal{F})$ of the style target $x_s$ The rather time-consuming optimization is turned into a simple forward with trained balancing weight, making real-time style transfer possible. However, the content and style information are still highly entangled as the style representation is the gram matrix of the content representation.

As pointed out by \cite{demystify}, this line of work could all be regarded as a domain adaption problem by minimizing the maximum mean discrepancy (MMD) with a specific kernel:

\begin{align*} 
& \operatorname{MMD}^{2}[\mathcal{X}, \mathcal{Y}] \\
=&\left\|\mathbf{E}_{x}[\phi(\mathbf{x})]-\mathbf{E}_{y}[\phi(\mathbf{y})]\right\|^{2} \\
\approx & \frac{1}{n^{2}} \sum_{i=1}^{n} \sum_{i^{\prime}=1}^{n} \phi\left(\mathbf{x}_{i}\right)^{T} \phi\left(\mathbf{x}_{i^{\prime}}\right)+\frac{1}{m^{2}} \sum_{j=1}^{m} \sum_{j^{\prime}=1}^{m} \phi\left(\mathbf{y}_{j}\right)^{T} \phi\left(\mathbf{y}_{j^{\prime}}\right) \\
&-\frac{2}{n m} \sum_{i=1}^{n} \sum_{j=1}^{m} \phi\left(\mathbf{x}_{i}\right)^{T} \phi\left(\mathbf{y}_{j}\right) \\
\end{align*}

Then the content is adapted by minimizing the MMD between the $\mathbf{x}_c$'s feature map domain and the original image $\mathbf{x}$'s feature map domain with $k(\mathbf{x},\mathbf{y}) = \phi(\mathbf{x})^T\phi(\mathbf{y}) = \mathbf{x}^T\mathbf{y}$, i.e. the linear kernel, and the style is adapted by minimizing the MMD between the $\mathbf{x}_s$'s feature map domain and the original image $\mathbf{x}$'s feature map domain with $k(\mathbf{x},\mathbf{y}) = \phi(\mathbf{x})^T \phi(\mathbf{y}) = (\mathbf{x}^T\mathbf{y})^2$, i.e. the second-order polynomial kernel.

\subsection{GAN-based Style Transfer}
In a recent work by \cite{cycle-gan}, unpaired samples ${\mathbf{x},\mathbf{y}}$ of two domains $\mathcal{X}, \mathcal{Y}$ are used to train two translators that can translate samples from one domain to samples of another domain that are indistinguishable to a discriminator of that domain. Moreover, in order to prevent arbitrary mapping, they exploited the fact that the translation should be cycle consistent: 
\[ F(G(\mathbf{x})) \approx \mathbf{x}, \quad G(F(\mathbf{y})) \approx \mathbf{y} \]

$F$ and $G$ should approximately be the inverse of each other and both of them are bijections. Combining this cycle consistency constraint and the adversarial loss gave outstanding results in tasks including image style transfer. Specifically, their objective could be viewed as learning two auto-encoder $F \circ G: \mathcal{X} \to \mathcal{X}$ and $G \circ F: \mathcal{Y} \to \mathcal{Y}$ where the the intermediate representation is precisely the translation of the image into another domain.

However, with this line of work, we can only deal with two styles at a time, and there is no way to specify the strength of style or to interpolate between different styles.

\subsection{Disentangled Learning}
Disentangled learning has been proposed in multiple works. \citet{disentangled_lecun} separated the variability into two components: inter-class variability (or specified variability) $\mathbf{s}$ and inner-class variability (or unspecified variability) $\mathbf{z}$. In order to prevent the information from all flowing from $\mathbf{z}$, we could, conditioned on a supervised setting, enforce the reconstruction of two identically labelled instance to be faithfully reconstructed by only swapping the unspecified variability $\mathbf{s}$. More specifically, this idea could be exploited in video setting where temporal content (i.e.: the subject/background of the video) consistency is given, and only the pose of the subject would change (\citet{disentangled_video}). We aim to disentangle the hidden representation of images into content and style. However, as number of consistent-styled images are limited, we will augment the data set by running some aforementioned neural style transfer methods on COCO 2014 train (\cite{COCO})

\section{Content-Style Disentanglement}

\subsection{Architecture}
\begin{figure}[h]
    \centering
    \includegraphics[width=\linewidth]{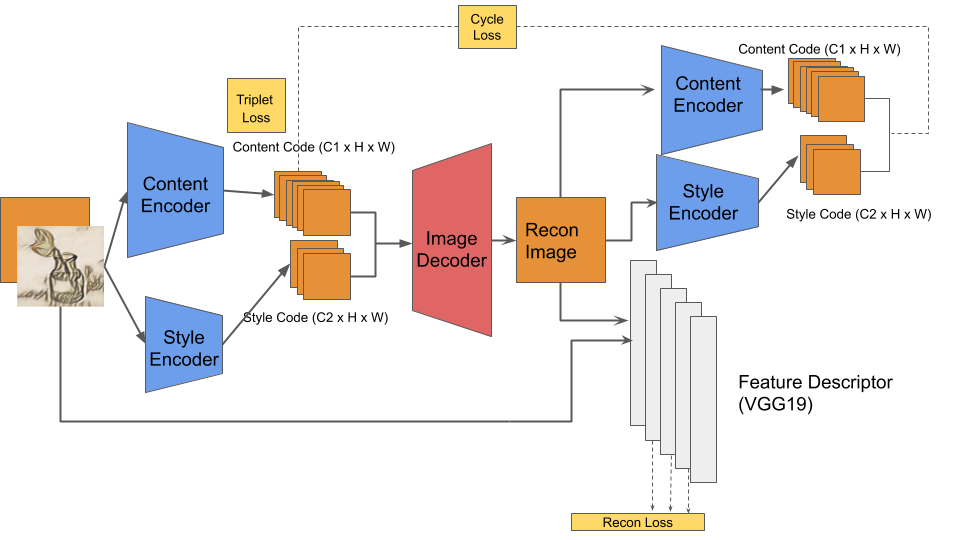}
    \caption{Model Structure}
    \label{fig:model}
\end{figure}

\begin{figure}[h]
    \centering
    \includegraphics[width=5cm]{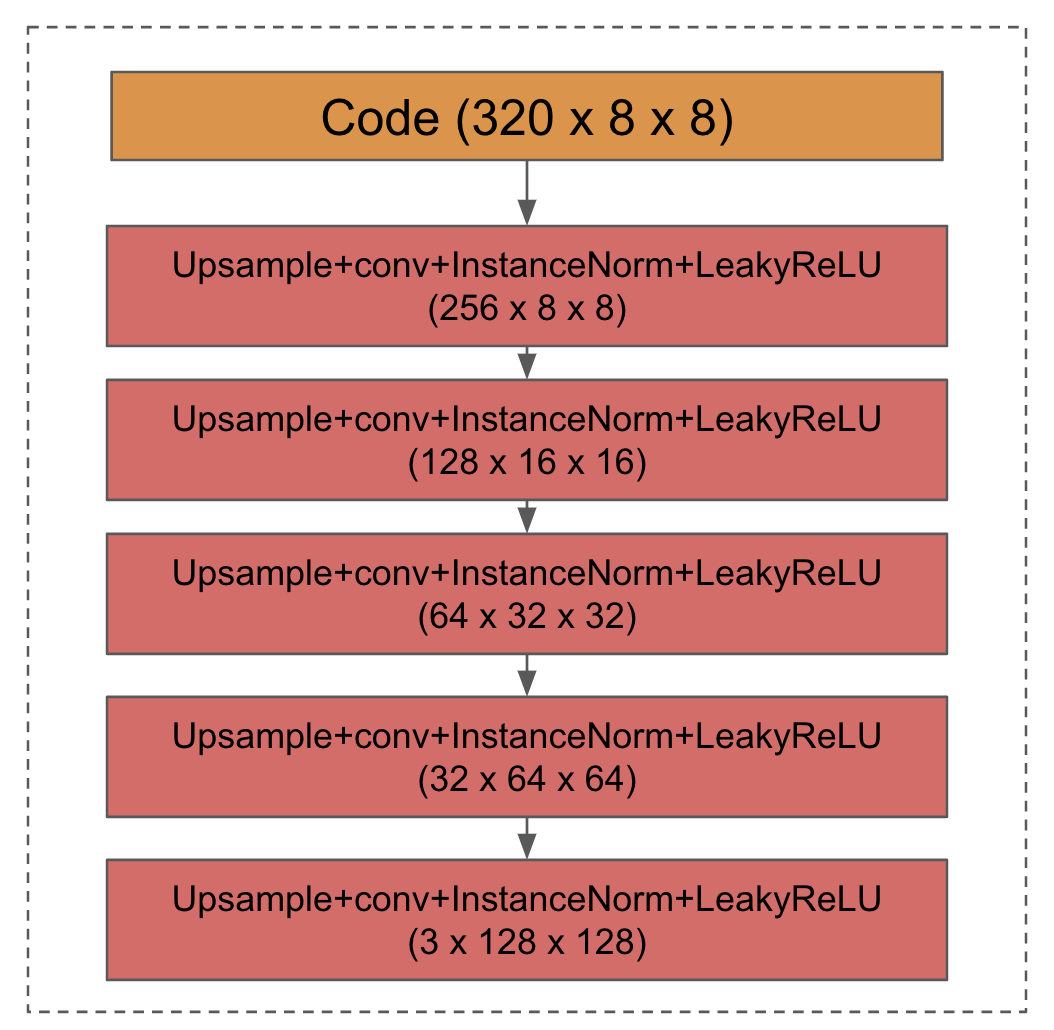} 
    \includegraphics[width=5cm]{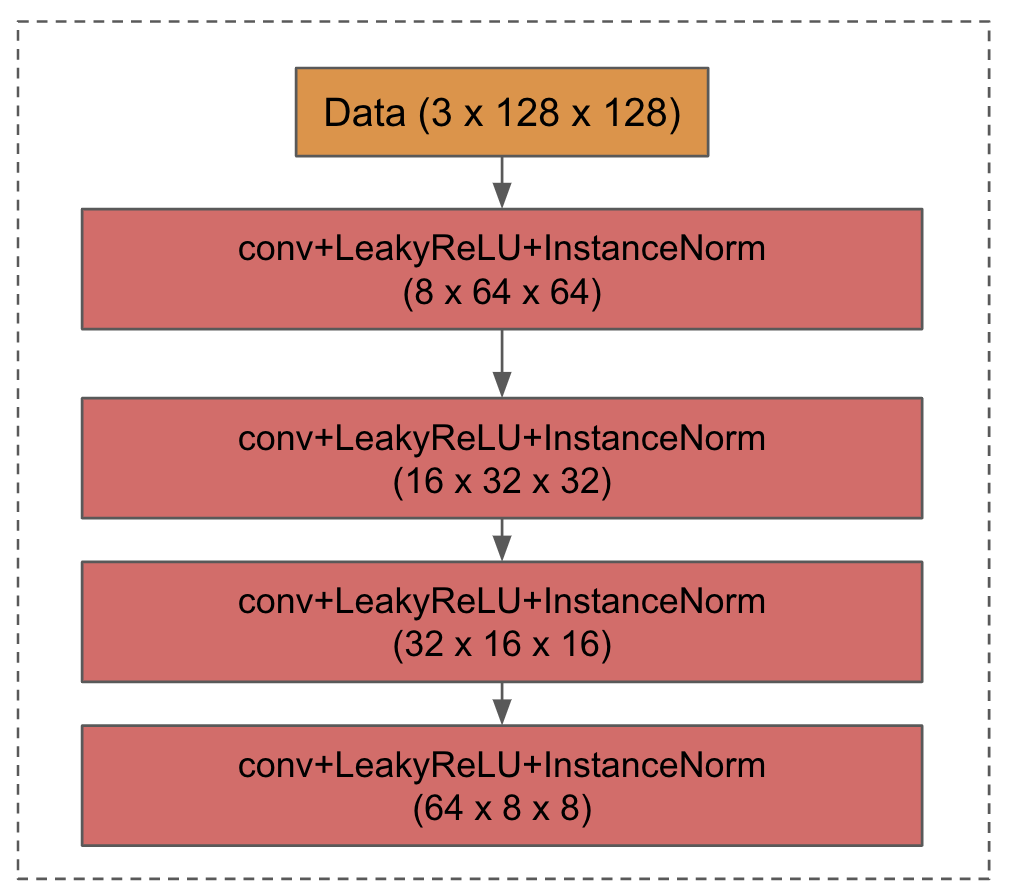} 
    \caption{Decoder (left) and Style Encoder (right)}%
    \label{fig:network_structure}%
\end{figure}

We propose a deep learning framework approach that disentangles style and content via a "Encoder-Decoder" style generative neural network model. We encode style and content using two different convolutional encoders. We then concatenate the two hidden codes and pass it through a single decoder network. The detailed structure is shown in Fig \ref{fig:model}. Our content encoder is a pretrained resnet34 proposed by \cite{he2016deep}, and our style encoder is a relatively shallow convolutional neural network. The Decoder follows the structure introduced by \cite{DFCVAE}. 
 The detailed structure of style encoder and the decoder is shown in Fig \ref{fig:network_structure}. We use Nearest Neighbor Sampling layer instead of Strided Convolution and LeakyReLU instead of ReLU to prevent checkerboard effect (\cite{odena2016deconvolution}). Additionally, we apply Instance Normalization proposed by \cite{Ulyanov2016InstanceNT} to improve time efficiency and image quality. 

Inspired by \cite{cycle-gan}, our model also utilizes transitivity of CNN to supervise training. As shown on Fig. \ref{fig:model}, the reconstructed image is again passed through the encoder, from which a cycle loss is incurred. We discuss the implementation details in section \ref{cycle-loss}.
\subsection{Reconstruction Loss}
Intuitively, reconstructed images should stay close to original images in $l_2$ distance. However, as we experiment with our proposed model above, we discover that $l_2$ loss between reconstructed tensor and original image tensor does not sufficiently capture structural information about the image, and results in extremely blurry reconstructed images. To avoid this issue, we apply a "feature descriptor" to encourage resemblance of reconstructed image and  original image in feature representation. Reconstruction loss consists of a content loss and a style loss. Similar to \cite{Johnson2016PerceptualLF}, content loss $L_{recon_c}$ computes \textit{mean square error (MSE)} of high-level feature maps, and style loss $L_{recon_s}$ computes \textit{MSE} of gram matrices of lower-level feature maps. By denoting descriptor network as $\phi$, the output of layer $j$ as $\phi_j$ of size $(C_j, H_j, W_j)$, and Gram matrix transformation as $G_j(\cdot)$, which output a tensor of size $(C_j, C_j)$, we have the reconstruction losses as follows:
\begin{align*}
    L_{recon_s}^{\phi_j}(\mathbf{x}, \mathbf{x}') &= \frac{1}{C_{j} H_{j} W_{j}} 
    ||G(\phi_{j}(\mathbf{x}))  - G(\phi_{j}(\mathbf{x}'))||_2^2 \\
    L_{recon_s}^\phi(X, \mathcal{E}, \mathcal{D}) &= \expt_{\mathbf{x} \sim \mathcal{X}, \mathbf{x}' \sim \mathcal{D(E(X))}}[\sum_j L^{\phi_j}_{recon_s}(\mathbf{x}, \mathbf{x}')]
    \\
    L_{recon_c}^{\phi_j}(\mathbf{x}, \mathbf{x}') &= \frac{1}{C_{j} H_{j} W_{j}} 
    ||\phi_{j}(\mathbf{x})  - \phi_{j}(\mathbf{x}')||_2^2 \\
    L_{recon_c}^\phi(X, \mathcal{E}, \mathcal{D}) &= \expt_{\mathbf{x} \sim \mathcal{X}, \mathbf{x}' \sim \mathcal{D(E(X))}}[\sum_j L^{\phi_j}_{recon_c}(\mathbf{x}, \mathbf{x}')]
\end{align*}

As for the choice of pretrained descriptor model, we use VGG16 proposed by \cite{simonyan2014very}. The style loss $L_{recon_s}$ is calculated from the output of layer \verb|relu1_2, relu2_2, relu3_3, relu4_3| and the content loss $L_{recon_c}$ is calculated from the output of layer \verb|relu3_1, relu4_1, relu5_1|. The intuition behind such construction of loss is that the style is captured by low level features and the content is defined by high level features in a image recognition system. (\cite{ghiasi2017exploring})
\subsection{Triplet Loss} \label{subsec:triplet}
We here refer to both \texttt{content} and \texttt{style} as "\texttt{property}", and refer to images of the same \texttt{property} (either \texttt{content} or \texttt{style}, depending on the context)  as belonging to the same "\texttt{class}". Then based on our assumption, the property code should be close for images of the same \texttt{class} and far apart for images of different \texttt{classes}. We provide a definition of \textbf{being of the same content or style} in section \ref{sec:dataset}. With this setup, we impose triplet loss in a similar manner to \cite{schroff2015facenet} and use $l_2$ distance as distance measure. Denote our \texttt{property} encoding network as $\mathcal{E}(\cdot)$. Then for any triplet ($\mathbf{x}_A, \mathbf{x}_P, \mathbf{x}_N$), where $\mathbf{x}_A$ and $\mathbf{x}_P$ share the same \texttt{property} and $\mathbf{x}_A$ and $\mathbf{x}_N$ have different ones, the triplet loss $L_{tri}$ is defined as the following:
\begin{align*}
    L_{tri}(\data_A, \data_P, \data_N) &= 
    \max(0,  
    \norm{\encoder(\data_A) - \encoder(\data_P)}_2^2 
    - \norm{\encoder(\data_A) - \encoder(\data_N)}_2^2 + \alpha_{margin}) \\
    L_{tri}(\mathcal{X}, \mathcal{E}) &= 
    \expt_{\data_A, \data_P, \data_N \sim \mathcal{X},
        \encoder(\data_P)=\encoder(\data_A), 
        \encoder(\data_N)\neq\encoder(\data_A)}
    [ L_{tri}(\mathbf{x}_A, \mathbf{x}_P, \mathbf{x}_N)]
\end{align*}
where $\alpha_{margin}$ is a hyper parameter represents the margin of distinction. We provide a detailed explanation of the sampling process we use to impose this loss in section \ref{sec:dataset}. 

\subsection{Cycle-Consistency Loss} \label{cycle-loss}
In order to prevent reconstructing an image that however has no relationship with the \texttt{content}/\texttt{style} code it is generated from, we want the hidden code to be cycle-consistent through the encoder-decoder-encoder cycle, i.e: if we pass the reconstructed image through both encoders again we would get exactly the codes it is generated from:
\begin{gather*}
\forall \mathbf{x} \in \mathcal{X}, \; 
\mathbf{e}_{\mathbf{x}, c} = \mathcal{E}_c(\mathbf{x}), \; 
\mathbf{e}_{\mathbf{x}, s} = \mathcal{E}_s(\mathbf{x}), \; \\
\mathcal{E}_c(\mathcal{D}(\mathbf{e}_{\mathbf{x},c}, \mathbf{e}_{\mathbf{x},s})) \approx \mathbf{e}_{\mathbf{x},c}, \;
\mathcal{E}_s(\mathcal{D}(\mathbf{e}_{\mathbf{x},c}, \mathbf{e}_{\mathbf{x},s})) \approx \mathbf{e}_{\mathbf{x},s}
\end{gather*}

This is similar to the forward cycle-consistency loss introduced by \cite{cycle-gan}. However, we apply our constraint on the encoded feature, whereas they add the penalty on cyclically generated samples directly. For the original image and the reconstructed image $\mathbf{x}, \mathbf{x}' \in \mathbb{R}^{C \times H \times W}$:
\begin{align*}
    L_{cycle}(\data, \data') &= 
    \frac{1}{CHW}\norm{\encoder_c(\data)-\encoder_c(\data')}_2^2 + \frac{1}{CHW}\norm{\encoder_s(\data)-\encoder_s(\data')}_2^2 \\
    L_{cycle}(\mathcal{X}, \encoder, \decoder) &= 
    \expt_{\data \sim \mathcal{X}, \data' \sim \decoder(\encoder(\mathcal{X}))}[L_{cycle}(\data, \data')]
\end{align*}

\subsection{Full Loss Function}
Finally, the full loss function is:
\begin{align*}
    L(\mathcal{X}, \mathcal{E}_c, \mathcal{E}_s, \mathcal{D}, \phi) &= \lambda_{s} L_{recon_s}^{\phi}(\mathcal{X}, \mathcal{E}, \mathcal{D}) + \lambda_c L_{recon_c}^{\phi}(\mathcal{X}, \mathcal{E}, \mathcal{D}) \\
    &\; + \lambda_{tri} L_{tri}(\mathcal{X}, \mathcal{E})  + \lambda_{cycle} L_{cycle}(\mathcal{X}, \mathcal{E}, \mathcal{D})
\end{align*}

\section{Dataset}\label{sec:dataset}
Training the model we proposed in the project requires a dataset that contains images with same content and different styles as well as images with same style but different contents. To our knowledge, no such dataset exists, with both aligned contents and aligned styles. Therefore we generate our own dataset by applying a set of styles to a set of non-stylistic images. We adopt the style transfer process proposed by  \cite{Johnson2016PerceptualLF} and improved by \cite{dumoulin2017learned}, in which they utilize learned weights in a CNN as basis for content and style loss. We use 32 style models to stylize 5000 Microsoft COCO images into a dataset of size $32 \times 5000$. In this project, we refer to images transferred by different style models from the same original COCO image as having the same \texttt{content} and images transferred by the same style model from different original COCO images as having the same \texttt{style}.
\begin{figure}[H]
 \centering
    \includegraphics[width=12cm, height=3cm]{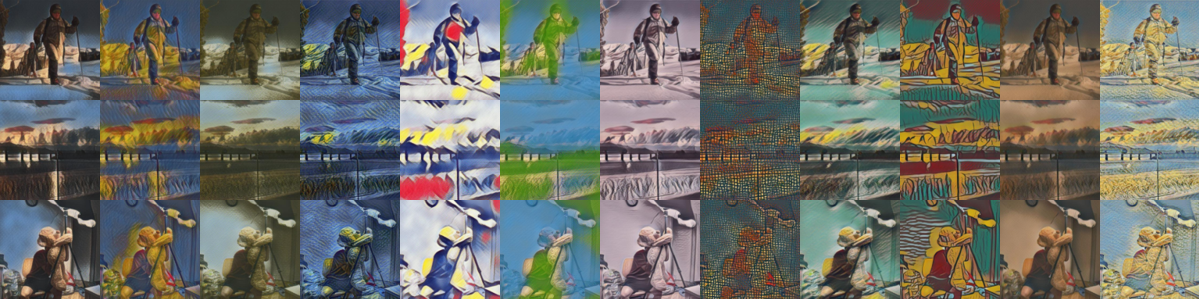} 
    \includegraphics[width=1cm, height=3cm]{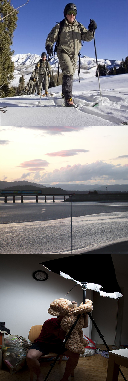} 
    \captionsetup{justification=centering}
    \caption{\textbf{Dataset}, 
    \\ The right most column is the original image from COCO \cite{COCO}.
    }%
    \label{fig:dataset}%
\end{figure}

In order to impose the triplet loss in section \ref{subsec:triplet}, we apply the following sampling process. For each triple:

\begin{itemize}
 \setlength\itemsep{-1em}
    \item uniformly sample two contents without replacement \texttt{content-1}, \texttt{content-2} \\
    \item uniformly sample one style \texttt{style} \\
\end{itemize}
Then the \texttt{content} triplet and \texttt{style} triplet in the order of $\{\mathbf{x}_A, \mathbf{x}_P, \mathbf{x}_N\}$ would be:
\begin{itemize}
\item $\{(\texttt{content-1}, \texttt{style}), (\texttt{content-1}, \emptyset), (\texttt{content-2}, \texttt{style})\}$ 
\item $\{(\texttt{content-1}, \texttt{style}), (\texttt{content-2}, \texttt{style}), (\texttt{content-1}, \emptyset)\}$
\end{itemize}
where $\emptyset$ means original image in COCO with no style applied.


\section{Results}
\subsection{Training Detail}
We implement all our models and training/inference procedure in Pytorch. For all models (encoders $\encoder_c$ and $\encoder_s$, decoder $\decoder$), we use Adam for optimization (\cite{kingma2014adam}) with a initial learning rate of \verb 5e-4  and decreased by a factor of \verb 0.2  for every \verb 30 epochs. The reconstruction weights for content/style ($\lambda_s/ \lambda_c$) are set to \verb|1.0|, the triplet weight $\lambda_{tri}$ is also set to \verb|1.0|, the cycle weight $\lambda_{cycle}$ is set to \verb|0.01|. We use different margin for content/style triplet loss to improve performances. The content margin is set to \verb|1.0| whereas the style margin is set to \verb|5.0|. The batch-size is set to 300 for each input of size \verb|3x128x128|. All experiments are done on AWS instances. 


\subsection{Reconstruction}
\begin{figure}
    \centering
    \includegraphics[width=\linewidth]{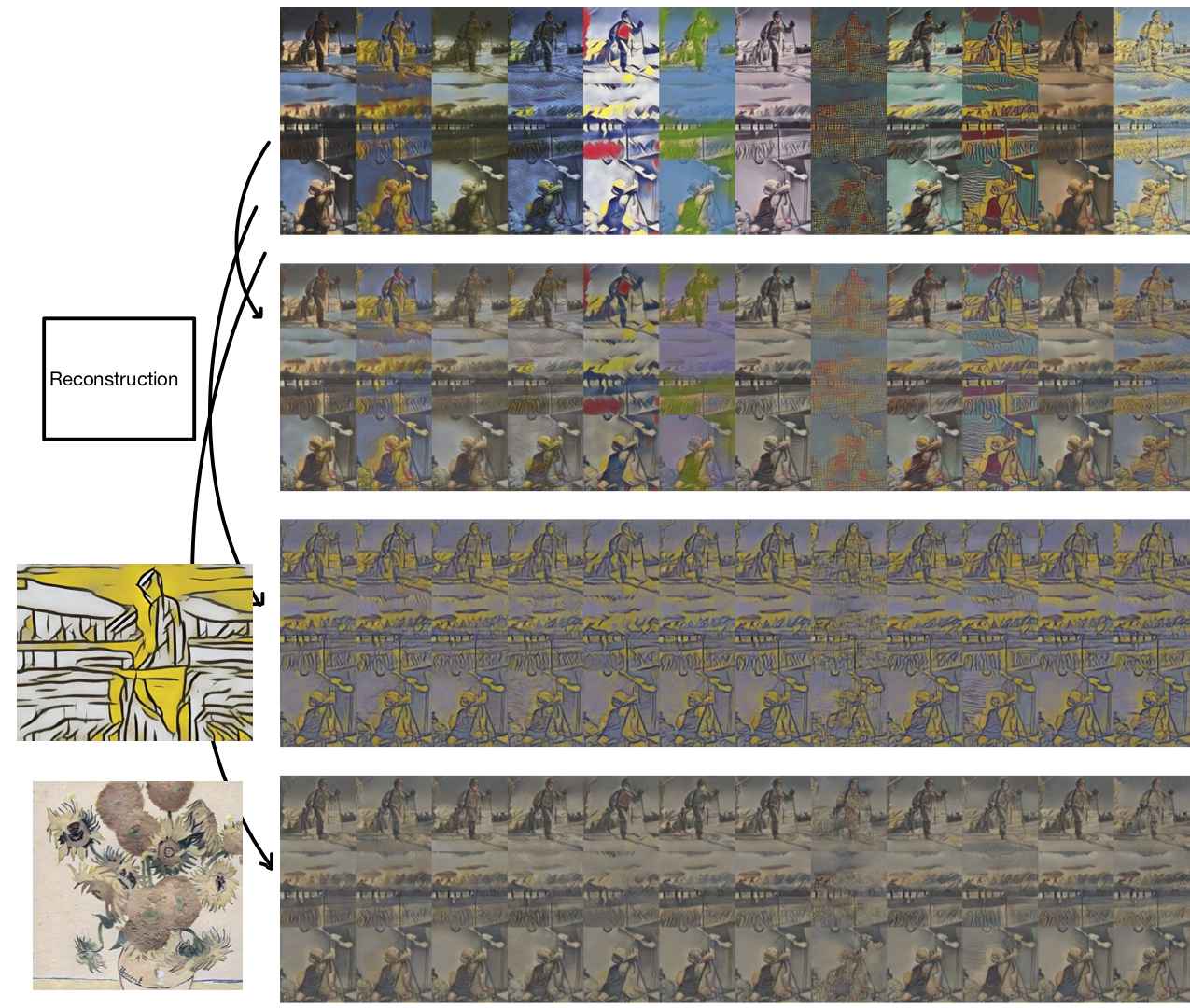}
    \captionsetup{justification=centering}
    \caption{\textbf{Reconstruction, Style Extrapolation}  \\
    The uppermost block lists the original images from our generated dataset.
    The second block lists our reconstruction result in the corresponding position as in the uppermost block. 
    The third block are our style transfer result with previously seen target style image listed on the left. 
    The four block has an unseen target style image listed on the left}
    \label{fig:707pic}
\end{figure}

Fig \ref{fig:707pic}'s second block shows our reconstruction result. Even for content blurry stylized images (e.g. 5th from right), our model is able to reconstruct both content/style.
\begin{figure}[h]
    \centering
    \includegraphics[width=\linewidth]{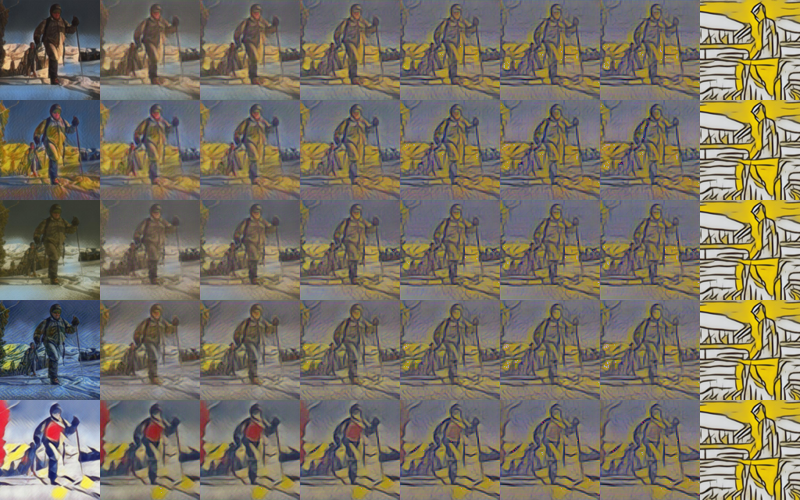}
    \captionsetup{justification=centering}
    \caption{\textbf{Interpolation} \\ Images in the left most column have a certain style and the image in the right most column has another distinct style. A series of images from left to right shows the result generated from interpolating these two styles with increasing weight of the right most column's style.}
    \label{fig:interpolation}
\end{figure}
\subsection{Style Disentangle}
\begin{figure}[h]
    \centering
    \includegraphics[width=8cm]{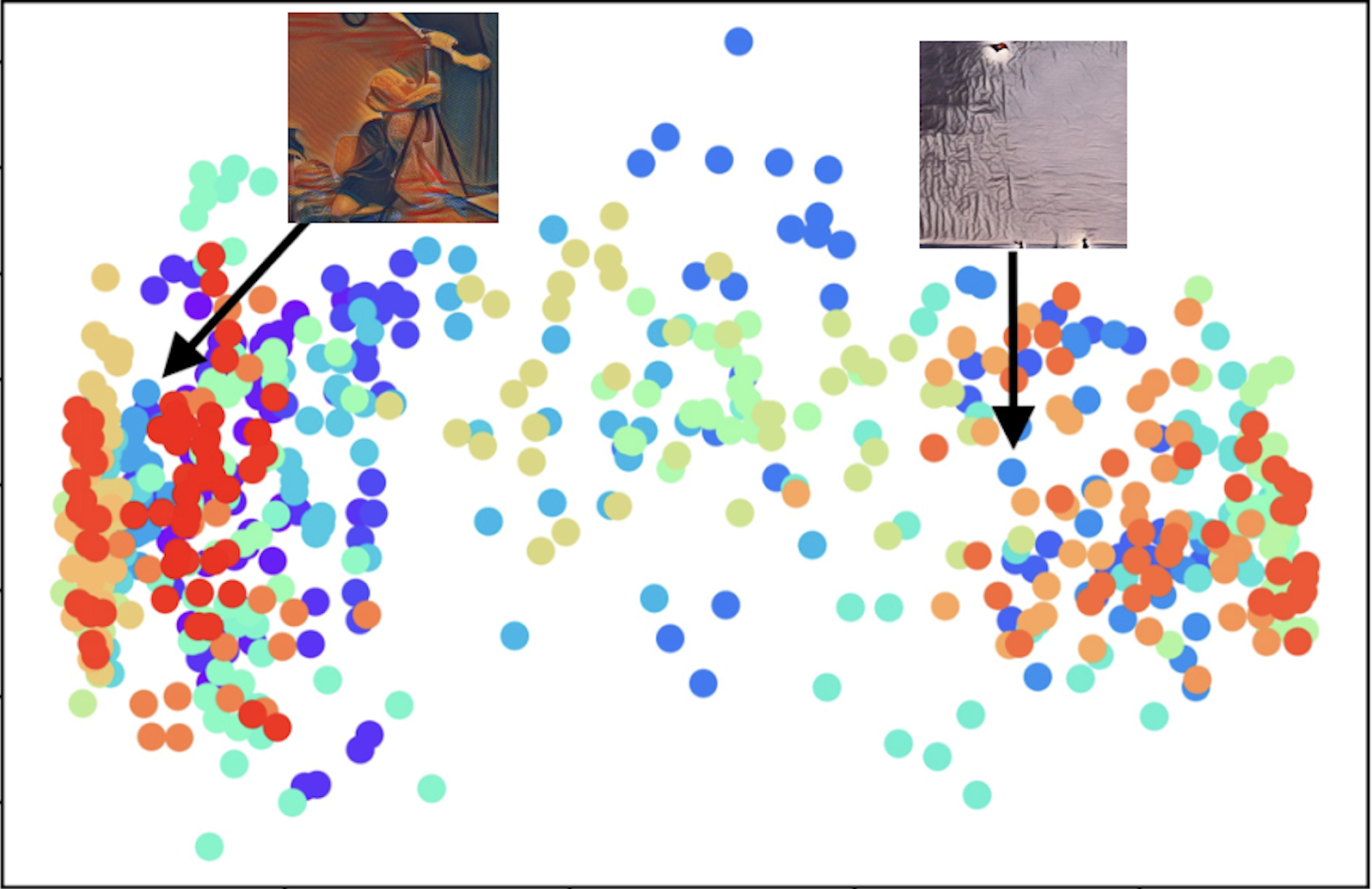}
    \caption{PCA result of style encoding}
    \label{fig:PCA}
\end{figure}

In order to show the effectiveness of our proposed disentangle method, the style code from style encoder is extracted and visualized. The 2-D plot Fig. \ref{fig:PCA} (Computed using PCA \cite{PCA}) shows our disentangle results. Each point represent a single image's style code, different color represent different styles. The cluster type structure is observed in the plot, two sample images are given as references.

\subsection{Style Extrapolation and Transfer}
Previous neural style transfer work done by \cite{cycle-gan} or \cite{Johnson2016PerceptualLF} only focus on two styles at a time, that is, their model is only capable of transforming one image to another style. Our proposed disentangle approach allows the transformation to any style possible. In this section, we transform image in our dataset into two different target styles: one is an observed style in our dataset and the other one is an unseen style. Shown in Fig \ref{fig:707pic}, the third block represent the transformation to an observed style, the result image has thick lines and yellow like color which is consistent with the human perception of "style" in target image. The forth block is the transfer result of an unseen image, as a result, our model is able to accurately capture the tone of the target image. In all the experiments, even if the input image is blurry, our model is able to extract sufficient content information to construct a high-resolution output.

\subsection{Style Interpolation}
As shown in Fig \ref{fig:interpolation}, we investigate the convex linear interpolation of our style code. Let two distinct style code be $C_{S_A}$ and $C_{S_B}$, The interpolation is defined by linear transformation:
\begin{align*}
    C_S = (1 - \alpha)C_{S_A} + \alpha C_{S_B}
\end{align*}
Where $\alpha$ ranges from $[0, 0.1, 0.3, 0.5, 0.7, 0.9, 1.0]$. The new style code is then concatenated with the original content code $C_c$ and fed into the decoder to generate new stylized images. From left to right, we can observe a smooth transformation from one style to another: red and green gradually diminish and yellow become more and more obvious, and that detailed edge structures (e.g. third and forth rows) are replaced by thicker strokes.

\subsection{Structure Exploration}
For our deep learning approach, we explored different structures for our model. For feature descriptor, we tested with both VGG16 \cite{simonyan2014very} and resnet34 \cite{he2016deep}. For decoder structure, we experiment with three different decoder structures: DCGAN \cite{radford2015unsupervised} style, Pixelvae \cite{pixelvae} style and DFC VAE \cite{DFCVAE} style. For encoder structure, we experiment with two different types of encoder: Conv-ReLU-Pool based encoder and DFC VAE styled encoder. We also explored 2-d and 3-d style/content code structure. In all our attempts, we find that the best combination is to use DFC VAE styled encoder and decoder together with 3-d feature and VGG16 as descriptor. All results shown above are based on this architecture.

\section{Discussion}
One of the major obstacles we encountered during the completion of this project is the lack of quality dataset. The dataset we generated contains 32 styles that depends on pretrained style models that vary in quality. Moreover, some of the style models are too strong that the content of transformed images are extremely hard to recover, even by human eyes. In hindsight, our proposed model would benefit significantly from a more thoughtfully and carefully crafted dataset. Although we have overcome several difficulties to achieve the performance stated above, there is still much left to be desired. For example, both the reconstructed and stylized images seem to be dimmer than ground truth.

On a higher level, our ground truth definition of style depends specifically on the stylized images generated from style models mentioned above, and we assume that this "style" aligns with human's conceptual notion of "style". The validity of such assumption remains an open question. 

\newpage
\bibliography{reference.bib}

\end{document}